\documentclass[letterpaper, 10 pt, conference]{ieeeconf}  

\IEEEoverridecommandlockouts                              

\usepackage{graphicx}
\usepackage{color}
\usepackage{cite}
\usepackage[hidelinks=true,bookmarks=false]{hyperref}
\usepackage{multirow}
\usepackage{enumerate}
\usepackage{bm}
\usepackage{subfigure}
\usepackage{booktabs}
\usepackage{blindtext}
\usepackage[T1]{fontenc}
\usepackage{mathptmx} 
\DeclareMathAlphabet{\mathcal}{OMS}{cmsy}{m}{n}

\usepackage{times} 
\usepackage{amsmath} 
\usepackage{amssymb}  
\usepackage[ruled,linesnumbered]{algorithm2e}
\usepackage{url}
\usepackage{bbm}
\usepackage{booktabs}
\usepackage{upgreek}

\title{\LARGE \bf
Autonomous Navigation through intersections with Graph Convolutional Networks and Conditional Imitation Learning for Self-driving Cars
}

\author{Xiaodong~Mei,
        Yuxiang~Sun,
        Yuying~Chen,
        Congcong~Liu,
        and~Ming~Liu,~\IEEEmembership{Senior Member,~IEEE}

\thanks{All authors are with The Hong Kong University of Science and Technology (email: {xmeiab, eeyxsun, ychenco, cliubh, eelium}@ust.hk).}
}

\begin{document}

\maketitle
\thispagestyle{empty}
\pagestyle{empty}

\begin{abstract}

In autonomous driving, navigation through unsignaled intersections with many traffic participants moving around is a challenging task. 
To provide a solution to this problem, we propose a novel branched network \texttt{G-CIL} for the navigation policy learning.
Specifically, we firstly represent such dynamic environments as graph-structured data and propose an effective strategy for edge definition to aggregate surrounding information for the ego-vehicle.
Then graph convolutional neural networks are used as the perception module to capture global and geometric features from the environment. 
To generate safe and efficient navigation policy, we further incorporate it with conditional imitation learning algorithm, to learn driving behaviors directly from expert demonstrations. 
Our proposed network is capable of handling a varying number of surrounding vehicles and generating optimal control actions (e.g., steering angle and throttle) according to the given high-level commands (e.g., turn left towards the global goal).
Evaluations on unsignaled intersections with various traffic density demonstrate that our end-to-end trainable neural network outperforms the baselines with higher success rate and shorter navigation time.


\end{abstract}

\section{Introduction}

\begin{figure}[tpb]
    \centering
    \setlength{\abovecaptionskip}{0pt} 
    \includegraphics[width = \columnwidth]{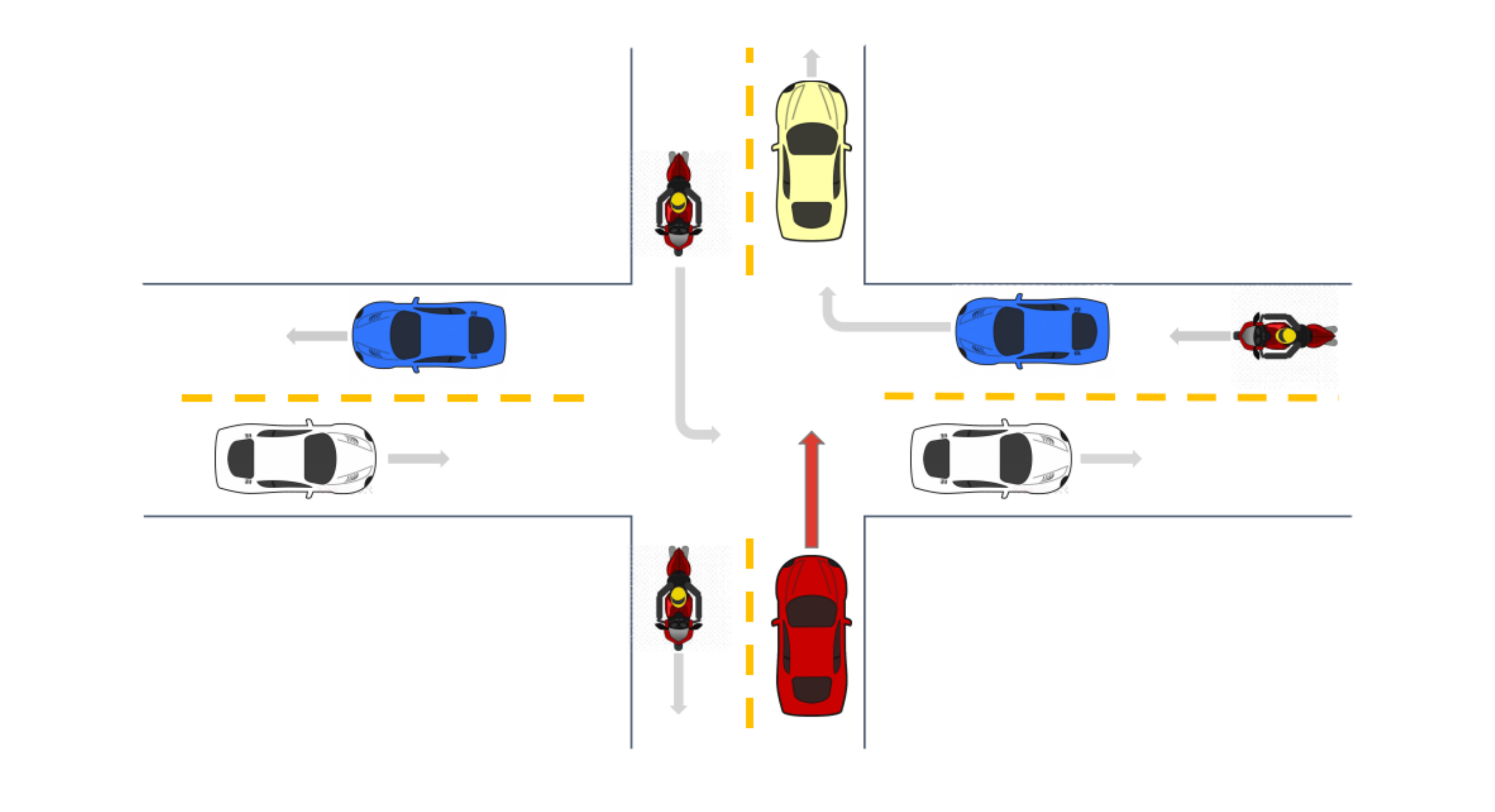}
    \caption{A typical unsignaled intersection with lots of surrounding traffic participants. Arrows in front of vehicles and bicycles refer to moving directions. The task of the ego-vehicle (red) is to drive through the intersection without collisions as far as possible.}
    \label{fig:1}
    \vspace{-0.01cm} 
\end{figure}

Navigation in dynamic traffic environments safely and efficiently is one of the most crucial and challenging problems for autonomous driving, such as driving through unsignaled intersections in urban scenarios. With a large number of traffic participants moving around and without any global scheduling, traffic junctions can be extremely chaotic, shown in Figure \ref{fig:1}. However, such dangerous situations are quite common, especially driving in countryside and residential districts. To handle it, autonomous vehicles are required to understand surrounding environments and generate the optimal navigation policy to reach the target destination with collision avoidance behaviors. 

Current works proposed to navigate intersections can be divided into \textit{rule-based} methods and \textit{learning-based} methods. 
Classical rule-based methods tend to model dynamic environmental information explicitly with complex or heuristic algorithms, like time-to-collision (TTC) \cite{van1993time}. 
The effectiveness of generated driving policy is heavily dependent on the accuracy of hand-crafted feature representations, which, however, limits those strategies into simple scenarios because of expensive modeling process in dense traffic. 
For the extension to complicated environments, learning-based methods are proposed to formulate the problem into the mapping task with deep neural networks: from the perceptual input to the control action. Several methods based on deep reinforcement learning (DRL) have been successfully applied into intersection handling \cite{isele2018navigating, bouton2019safe}.
However, as the environment complexity increases, sample efficiency problem of DRL leads to the exponential growth in policy learning difficulty. 
By contrast, imitation learning (IL) is more efficient to learn the policy in a supervised way, that is, to mimic the expert behaviors. As an variant, conditional imitation learning (CIL) \cite{Codevilla2018} algorithm enables to learn the navigation policy conditioned on additional high-level commands (e.g., turn left when approaching an intersection), which improves navigation performance for autonomous driving in complex urban scenarios. 

Although navigation policy learned with CIL is able to predict the optimal control action from raw sensory inputs, it is typical to incorporate the perception module into the system, to extract high-level feature representations previously for the better understanding of complicated scenes and generalization ability enhancement.
In previous works, states in dynamic environments represented by vector features are encoded with MLP or LSTM, which causes the unawareness of agent interactions and geometric information. 
However, dynamic environments with many traffic participants are natural to be represented by graphs, where nodes refer to different agents and edges are defined to represent neighborhood relationships.
Graph convolutional networks (GCNs) \cite{kipf2017semi, schlichtkrull2018modeling}, which is capable of handling graph-structured data and aggregating neighbors' information, completely captures global and dynamic features to learn more effective and robust environmental representations.

\begin{figure*}[t]
    \centering
    \setlength{\abovecaptionskip}{-0.5pt}
    \includegraphics[width = 1.9\columnwidth]{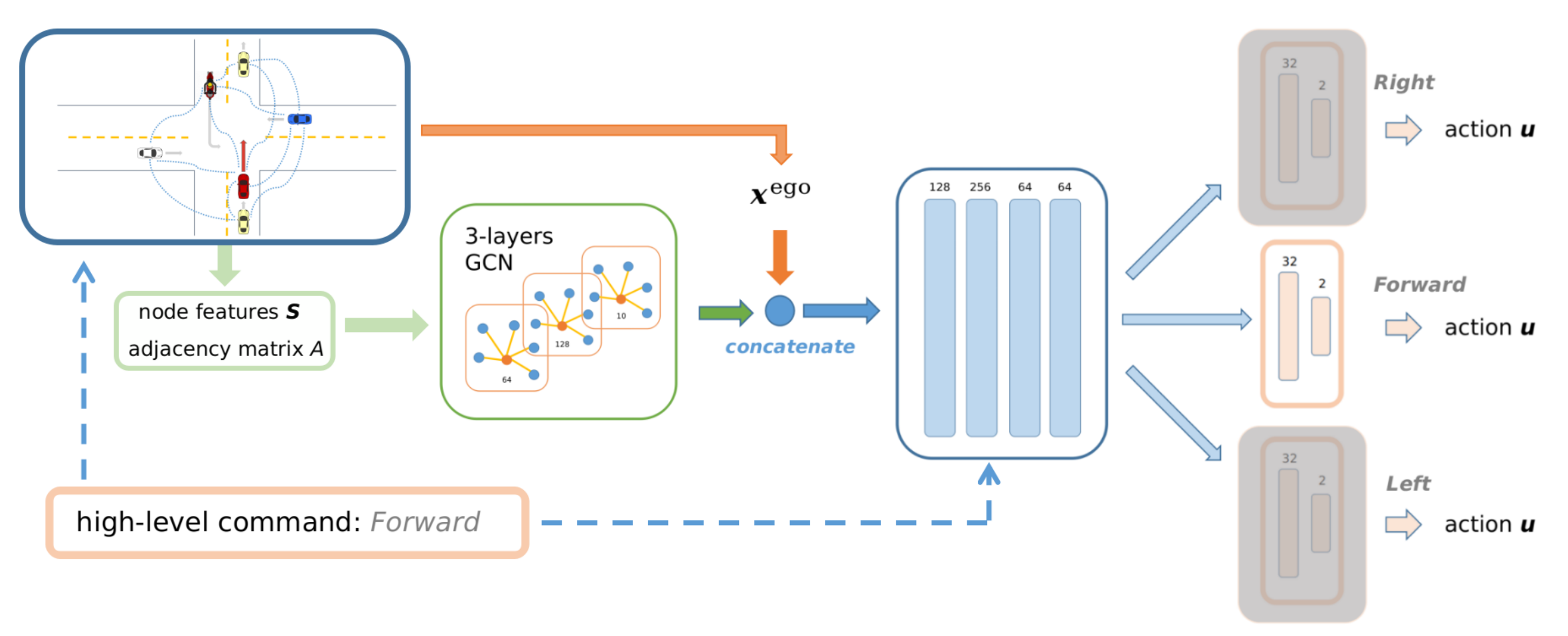}
    \caption{The proposed branched network \texttt{G-CIL}. All neural network layers represented with colored blocks are fully connected except for GCN layers and the numbers indicate individual output channel dimension. The branches in shadow mean \textit{non-working} since the high-level command is \textit{Forward} now. Dotted lines in the driving scene mean that the two vehicles are the defined neighbors.} 
    \label{fig:net}
    \vspace{-0.3cm}
\end{figure*}

In this paper, we propose a novel branched network \texttt{G-CIL} based on the combination of graph convolutional networks (GCNs) and conditional imitation learning (CIL) for autonomous navigation through unsignaled intersections. 
Environmental information is represented as graph-structured data and mapped into control actions for the ego-vehicle, directed by high-level commands. The main contributions of our paper are as follows:
\begin{itemize}

\item For navigating unsignaled intersections, we represent dynamic environments as graph-structure data and propose an effective edge definition strategy to capture surrounding information for the ego-vehicle. Ablation study shows that our strategy enhances the dynamic scene understanding ability and navigation performance. 
\item We design a command-conditional framework \texttt{G-CIL} to learn the navigation policy through unsignaled intersections for autonomous vehicles. We explore the navigation ability with comprehensive experiments and demonstrate that the driving policy is able to generate safe and efficient control actions to reach the target destination without collisions. 
\item We show that the proposed method has better generalization ability to unseen environments with higher success rate and shorter navigation time, compared with baseline methods.

\end{itemize}

\section{Related Work}

\subsection{Policy Learning in Dynamic Environments}

Since modeling dynamic environments explicitly and preciously are complicated and expensive for autonomous driving, researchers recently pay more attention to learning-based strategies instead of rule-based ones. With deep neural networks as non-linear function approximators, learning-based methods frame the navigation task as the mapping problem, that is, mapping the perceptual input into the optimal action. Several algorithms have been applied to navigation policy learning, such as deep reinforcement learning (DRL) \cite{cai2020high, tai2017virtual}, inverse reinforcement learning (IRL) \cite{kretzschmar2016socially}, and imitation learning (IL) \cite{cai2019vision, cai2020probabilistic, cai2020vtgnet, tai2018socially}. 

DRL algorithms learn the policy by trial-and-error, which is fully dependent on the experience collected through interactions with environments. 
There exist two main limitations for DRL methods. Firstly, the optimal policy learning tends to be extremely difficult in terms of large amounts of collected experience, which is known as the sample efficiency problem. Secondly, the reward function, which has a great influence on policy performance, is hard to design for complicated tasks. To address the reward engineering problem, inverse reinforcement learning (IRL) gives a possible solution to learn the reward from training data. Furthermore, imitation learning algorithms directly learn the optimal action from training data provided by the expert, which formulates the policy learning with the idea of supervised learning and greatly improves training efficiency in complex environments.

The main approach of imitation learning is also known as behavior cloning (BC), which aims to learn the optimal policy by mimicking an expert. 
Since driving behaviors rely not only on perceptual inputs but also the expert's internal state (e.g., the moving direction to reach the target destination), conditional imitation learning \cite{Codevilla2018} is proposed to take high-level control commands (e.g., \textit{Turn Left}) into consideration. Based on it, we build the branched network architecture conditioned on three high-level control commands, to predict more accurate control actions accordingly.

\subsection{Graph Representation Learning}

In the context of navigation through dynamic environments, previous works handle surrounding information with pooling \cite{chen2019crowd, deo2018convolutional}, maximum \cite{chen2017decentralized}, concatenate \cite{ding2019predicting} and sum operation \cite{zaheer2017deep} or the LSTM model \cite{huegle2019dynamic}, which lose the global and geometric information more or less. With the rapid development of graph convolutional networks (GCNs) and its variants \cite{kipf2017semi, schlichtkrull2018modeling, velivckovic2017graph}, researchers tend to represent dynamic driving scenarios with many vehicles as the graph-structured information \cite{huegle2020dynamic, hart2020graph}, instead of discrete vector features. GCN is efficient and effective to aggregate surrounding perceptual information on the basis of a set of neighbors defined by an adjacency matrix. And numerical values in the adjacency matrix can indicate the relationship of two agents with or without weights \cite{chen2020robot}. 

In terms of driving through unsignaled intersections, it is natural to consider the dynamic environment at the certain time step as a graph, where nodes refer to various traffic participants and edges are defined with respect to relative distances to indicate neighborhood relationships. We utilize the 3-layer GCN learn the perceptual representations, followed by the branched network to perform control action prediction conditioned on various high-level commands. 


\section{Methods}

We formulate the problem of navigation through unsignaled intersections as a conditional imitation learning problem with graph representations. We train graph convolutional networks (GCNs) as the perception module for the dynamic scene understanding, and the control module consisting of specialized sub-branches to output accurate control actions with respect to the specific command. The proposed branched network architecture and one action prediction step in \textit{Forward} task are shown in Figure \ref{fig:net}. For demonstration purposes, we use vehicles to represent surrounding agents, although in experiments surrounding agents can be bicycles. 


\subsection{Perception module: Graph Convolutional Network}

The fixed-size vectorial representation with MLP or LSTM is incapable of completely obtaining environmental information with varying number of surrounding vehicles in navigation task. Therefore, we introduce the graph representation to explicitly model the perceptual information. Typically, the graph is defined with a set of nodes together with individual node features formed as a matrix $\mathbf{S}$ and a set of edges represented by an adjacency matrix $\mathit{A}$. With the built graph, graph convolutional networks (GCNs) \cite{kipf2017semi} aggregate and transform information across nodes and output the learned feature vector for each node. For the multi-layer GCN, the layer-wise propagation rule is 

\begin{equation}
    \mathit{H^{(l+1)}} = \sigma(\mathit{A}\mathit{H^{(l)}\mathit{W}^{(l)}}),
\label{gcn-rule}
\end{equation}
where $\mathit{H}^{(0)} = \mathbf{S}$. $\sigma(\cdot)$ refers to the activation fuction and we use $\mathit{ReLU}(\cdot)$ in our three-layer GCN. $\mathit{W}^{(l)}$ is the trainable weight matrix in the $l^{th}$ layer.

For our navigation task, nodes refer to the ego-vehicle and surrounding vehicles, the total number of which is known as $\mathit{N}$. And the adjacency matrix $\mathit{A} \in \mathbb{R}^{N \times N}$ defines connections among these nodes. We propose the n-close vehicle connection strategy to model dynamic interactions in the scene, together with the self-connection of each node. For the node corresponding to the ego-vehicle, it connects to every surrounding vehicle in the scene with weighted edges; for other vehicles, each of them connects to the nearest three vehicles.
Specifically, $\mathit{a_{ij}} = 0$ means there is no connection between node $i$ and node $j$; otherwise, the value of $\mathit{a_{ij}}$ is the edge weight defined as follows
\begin{equation}
    \mathit{a_{ij}} = \mathit{e^{-d^{2}_{ij}/\alpha^2}},
    \mathit{i,j \in \{0,1,...,N-1\}}
\label{edge weight}
\end{equation}
where $\mathit{d_{ij}}$ is the relative distance between the corresponding nodes and the value is 0 for the self-connection. We choose $\alpha$ as 10 for the best performance. The constructed adjacency matrix $\mathit{A}$ is normalized where the sum of every row is one.

The input node features $\mathbf{S}$, which is also $\mathit{H^{(0)}}$, make the combination of information of the goal, the ego-vehicle and surrounding vehicles, represented as
\begin{equation}
    \begin{aligned}
        &\mathbf{s}_{i} = [\mathbf{x}^{ego}, \mathbf{x}_{i}],  \\
        &\mathbf{x}^{ego} = [d^{goal}, d_{x}^{goal}, d_{y}^{goal}, v_{err}, v_{x}, v_{y}], \\
        &\mathbf{x}_{i} = [d^{rel}_{i}, d_{xi}^{rel}, d_{yi}^{rel}, v^{rel}_{i}, v_{xi}^{rel}, v_{yi}^{rel}],
    \end{aligned}
\label{node feature}
\end{equation}
where $\mathit{i} \in \{0,1,...,\mathit{N-1}\}$. Each node feature includes the state of ego-vehicle $\mathbf{x}^{ego}$ and corresponding state $\mathbf{x}_{i}$ at current time step. $\mathbf{x}^{ego}$ consists of the relative distance between the ego-vehicle and the goal ($d^{goal}$), the relative distance along $\mathit{x}$ axis ($d_{x}^{goal}$) and $\mathit{y}$ axis ($d_{y}^{goal}$), the difference to the preferred velocity ($v_{err}$), and the velocity along $\mathit{x}$ axis ($v_{x}$) and $\mathit{y}$ axis ($v_{y}$). $\mathbf{x}_{i}$ consists of the individual state of every vehicle w.r.t ego-vehicle, including the relative distance ($d^{rel}_{i}$), the relative distance along $\mathit{x}$ axis ($d_{xi}^{rel}$) and $\mathit{y}$ axis ($d_{yi}^{rel}$), the relative velocity ($v^{rel}_{i}$), and the relative velocity along $\mathit{x}$ axis ($v_{xi}^{rel}$) and $\mathit{y}$ axis ($v_{yi}^{rel}$). 

For every time step $\mathit{t}$ in our navigation task, the three-layer GCN performs as the perception module, to take the node features $\mathbf{S}_t$ and the adjacency matrix $\mathit{A}_t$ as the input and output the ten dimensional feature for every node $\mathit{i}$. The output node feature corresponding to the ego-vehicle, combined with $\mathbf{x}^{ego}$ as the perception information $\mathbf{p}_t$, passes through the branched control module for action $\mathbf{u}_t$ prediction. 

\begin{figure*}[th]
    \centering
    \setlength{\abovecaptionskip}{-0.5pt}
    \includegraphics[width = 2.05\columnwidth]{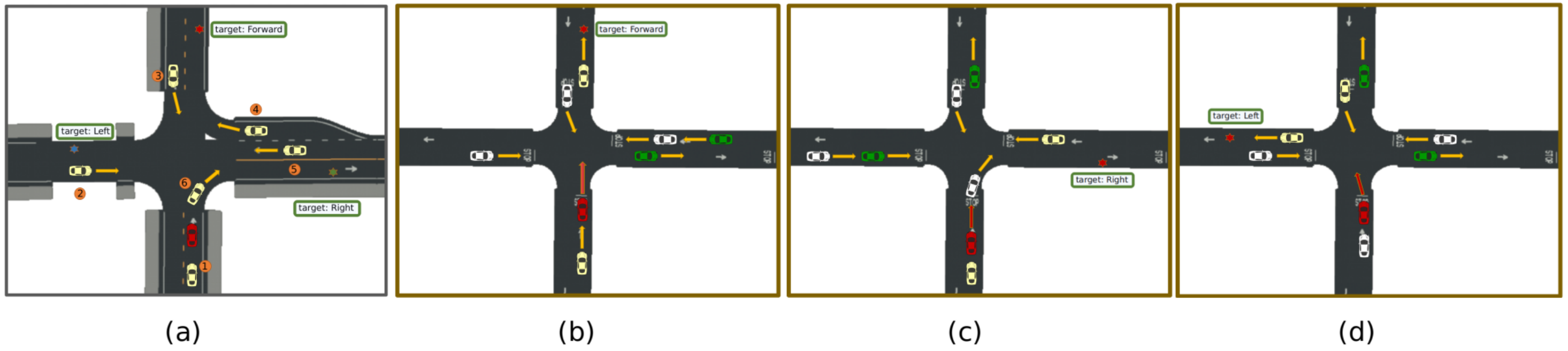}
    \caption{The training environment (a) and testing environments (b-d). Red one is the ego-vehicle. Target destinations are presented with stars. Arrows in orange color indicate moving directions. Note that we use the same type of vehicles to represent surrounding agents in the figure, but in experiments there might be various types of vehicles or bicycles. For environment (a),(c) and (d), the behind agent without the moving direction arrow always has the same target destination with the ego-vehicle. Moreover, relative distances shown in the figure are not the same as the values in experiments at the beginning, since the start position of every agent is sampled along the road.} 
    \label{fig:scene}
    \vspace{-0.3cm}
\end{figure*}

\subsection{Control module: Conditional Imitation Learning}

The ego-vehicle learns to generate correct actions by mimicking the expert in imitation learning. Different from the standard imitation learning, conditional limitation learning models the expert's internal state (e.g., the command $\mathbf{c}^{i}$: \textit{Turn left}) as well as the information $\mathbf{p}_t$ received by the ego-vehicle at the time step $t$. Following the branched network architecture in \cite{Codevilla2018}, we formulate the action prediction as 
\begin{equation}
    \mathit{f}(\mathbf{p}_t,  \mathbf{c}^{i}) = \mathit{U^i}(\mathit{g}(\mathbf{p}_t)),
\label{CIL}
\end{equation}
$\mathbf{c}^i \in \mathbf{C}$ refers to the control command w.r.t. the goal when driving through intersections, such as \textit{Forward}, \textit{Turn Left} and \textit{Turn Right}. The network branch $\mathit{U^i}$ learns correct actions according to different commands.

As is shown in Figure \ref{fig:net}, $\mathbf{p}_t$ firstly passes through four-layer fully connected neural networks with (128, 256, 64, 64) units respectively, and then is mapped with two-layer fully connected neural networks as the specialized branch into the two-dimensional action $\mathbf{u}_t$. The activation function of every layer is $\mathit{ReLU}(\cdot)$ expect for the last one, which is $\mathit{tanh}(\cdot)$ to limit the output range as [-1, 1]. The learned action $\mathbf{u}$ in our navigation task is the steering angle $\delta$ and throttle $\tau$.    




\subsection{Loss Function and Training Procedure}

With the expert action $\mathbf{u^\star} = (\delta^\star, \tau^\star)$, the loss function is defined as
\begin{equation}
    \mathit{l}(\mathbf{u},  \mathbf{u}^{\star}) = MSE(\delta, \delta^\star) + MSE(\tau, \tau^\star),
\label{loss}
\end{equation}

The proposed model is end-to-end trainable and all weights are shared expect for the branches. To learn the command-conditional policy, we construct three buffers to store the training data $\mathcal{D} = \{<(\mathbf{S}_t, \mathit{A}_t), \mathbf{u}_t>\}^{T}_{t=0}$.  
For every training step, minibatches of size 512 are sampled equally from different buffers to update GCN, four-layer MLP and the corresponding branch with Adam optimizer \cite{kingma2014adam}.

\begin{center}
    \begin{table*}[th]
    \newcommand{\tabincell}[2]{\begin{tabular}{@{}#1@{}}#2\end{tabular}}
    \newcommand{\NA}{---}
            \setlength{\abovecaptionskip}{5pt}
            \renewcommand{\arraystretch}{1.5}
            \caption{Quantitative Evaluation Results for Different Methods under Varied Environment Setups.}
            \label{result_table}
            \centering
            \setlength{\tabcolsep}{0.015\columnwidth}{
            \begin{tabular}{c c c c c c c c c c  c c c}
            \toprule
            \multirow{2}{*}{Setup}&
            \multirow{2}{*}{Methods}& 
            \multicolumn{4}{c}{{Success Rate (\%)}}&
            \multicolumn{4}{c}{{Collision Rate (\%)}}&
            \multicolumn{3}{c}{{Time ($s$)}}\\
            \cmidrule(lr){3-6} \cmidrule(lr){7-10} \cmidrule(lr){11-13}
            \multirow{1}{*}{}&
            \multirow{1}{*}{} 
            & {Forward} ($\uparrow$) & {Right} ($\Rsh$) & {Left} ($\Lsh$) & {AVG} & {Forward} ($\uparrow$) & {Right} ($\Rsh$) & {Left} ($\Lsh$) & {AVG} & {Forward} ($\uparrow$) & {Right} ($\Rsh$) & {Left} ($\Lsh$)\\
            \midrule
            \multirow{3}{*}{\tabincell{c}{N = 4}}&
            \texttt{G-CIL(ours)} & \textbf{78.57} & 57.14 & 75.71 & 70.47 & \textbf{21.43} & 42.86 & 24.29 & 29.53 & 15.44 & 10.71 & 11.86\\
            
            \multirow{3}{*}{}&  
            \texttt{NN-CIL} & 55.71 & \textbf{84.29} & 81.43 & \textbf{73.81} & 32.86 & \textbf{15.71} & 18.57 & \textbf{22.38} & 18.56 & 10.59 & 11.67\\
            \multirow{3}{*}{}&
            \texttt{Set-CIL} & 35.71 & 74.29 & \textbf{100.00} & 70.00 & 64.29 & 25.71& \textbf{0.00} & 30.00 & \textbf{14.88} & \textbf{10.50} & \textbf{11.38}\\
            \cline{2-13}
            
            \multirow{3}{*}{\tabincell{c}{N = 6}}&
            \texttt{G-CIL(ours)} & \textbf{77.14} & \textbf{65.71} & \textbf{61.43} & \textbf{68.09} & \textbf{22.86} & \textbf{34.29} & \textbf{38.57} & \textbf{31.91} & \textbf{15.38} & 10.89 & 11.86\\
            
            \multirow{3}{*}{}&  
            \texttt{NN-CIL} & 62.86 & 15.71 & 48.57 & 42.38 & 34.29 & 84.29 & 51.43 & 56.67 & 16.82 & 11.02 & 11.75\\
            
            \multirow{3}{*}{}&
            \texttt{Set-CIL} & 0.00 & 12.86 & 37.14 & 16.67 & 100.00 & 87.14 & 62.86 & 83.33 & NA & \textbf{10.75} &\textbf{11.51}\\
            
            \cline{2-13}
            
            \multirow{3}{*}{\tabincell{c}{N = 8}}&
            \texttt{G-CIL(ours)} & \textbf{52.86} & \textbf{44.29} & \textbf{45.71} & \textbf{47.52} & \textbf{47.14} & \textbf{55.71} & \textbf{54.29} & \textbf{52.38} & \textbf{15.47} & \textbf{10.89} & \textbf{11.89}\\
            \multirow{3}{*}{}&  
            \texttt{NN-CIL} & 42.86 & 28.57 & 38.57 & 36.67 & 55.71 & 71.43 & 61.43 & 62.86 & 17.54 & 11.11 & 11.94\\
            \multirow{3}{*}{}&
            \texttt{Set-CIL} & 0.00 & 0.00 & 32.86 & 10.95 & 100.00 & 100.00 & 67.14 & 89.05 & NA & NA & 12.21\\
            
            \bottomrule
            \end{tabular}}
    \label{table:1}
    \end{table*}
\end{center}

\begin{figure*}[h]
    \centering
    \setlength{\abovecaptionskip}{-0.5pt}
    \includegraphics[width = 2.0\columnwidth]{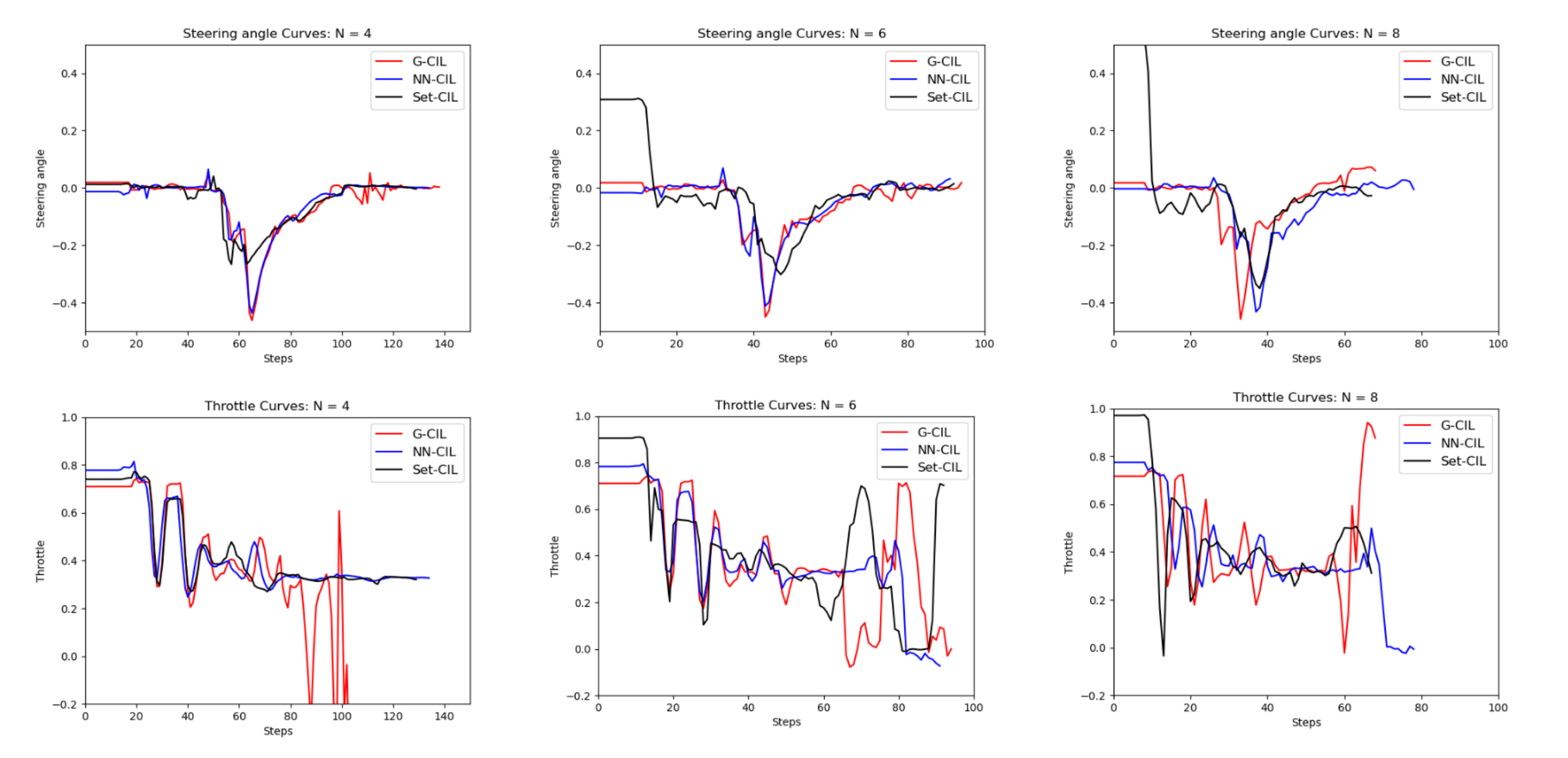}
    \caption{Steering angle and throttle curves of all three methods from successful trails for \textit{Turn Left} task with different environment setups. Red lines refer to output curves of \texttt{G-CIL}, blue lines refer to \texttt{NN-CIL} and black lines to \texttt{Set-CIL}.} 
    \label{fig:curve}
    \vspace{-0.3cm}
\end{figure*}

\section{Experiments}

In this section, we firstly introduce the training environment and testing environment setups. Then we introduce baselines and metrics for performance evaluation of our proposed \texttt{G-CIL}. We give the comprehensive analysis on both qualitative and quantitative aspects. Finally, we present the ablation study on our edge definition strategy.

\subsection{Environment Setups}

Training data collection is the most essential stage for imitation learning algorithms to obtain the optimal policy. We choose intersections with two-lane one-way roads in \textit{Town07} of the simulator CARLA \cite{dosovitskiy2017carla} for training and testing experiments, and road structures are shown in Figure \ref{fig:scene}. 

We spawn different number of vehicles for three training scenarios: in \textit{Forward} task, there are five vehicles (\textit{No.1, No.2, No.3, No.4 and No.5}); in \textit{Turn Left} task, there are only three vehicles  (\textit{No.2, No.3 and No.6}); in \textit{Turn Right} task, there are three ones (\textit{No.1, No.2, No.5}). All vehicles are controlled by \textit{BeheaviorAgent} algorithm in CARLA with the ignorance of ego-vehicle, traffic lights and traffic signs and their start positions are randomly sampled along lanes, so that velocities are also varying in different time steps. The designed environments ensure that every surrounding vehicle has overlapping trajectory with the ego-vehicle in future time steps, which means that ego-vehicle is unable to reach the target destination safely without intelligent control actions. We record both successful and failure cases from the expert to obtain the comprehensive training data set.

For evaluation, we design three scenarios of three traffic density levels (3, 5, 7 surrounding vehicles) with another intersection, which are totally nine kinds of scenarios. As is shown in Figure \ref{fig:scene} (b-d), three vehicles in white color are spawned in \texttt{easy} setups; then two yellow vehicles are added in \texttt{middle} setups; and in \texttt{hard} setups there are seven surrounding vehicles. 
Start positions and target destinations of both ego-vehicle and surrounding vehicles are sampled from a small range and different from the training experiments, for evaluations of the generalization ability. It is worth mentioning that in testing setups, trajectories of some vehicles will not overlap with the ego-vehicle, which is necessary because autonomous vehicle is required to distinguish whether nearby moving traffic participants influence it or not.       


\subsection{Comparative Analysis}

\subsubsection{Metrics} To evaluate the safety and effectiveness of our \texttt{G-CIL}, we utilize \textbf{success rate} (SR), \textbf{collision rate} (CR) and average \textbf{navigation time} in successful trials as evaluation metrics. \textit{Success} refers to reaching the target destination within a threshold distance and the equation is
\begin{equation}
    SR = \mathit{(N_{success}/N_{attempts}) \times 100\%},
\end{equation}
Similarly, collision rate is
\begin{equation}
    CR = \mathit{(N_{collision}/N_{attempts}) \times 100\%},
\end{equation}
Note that in our experiments, the sum result of success rate and collision rate might not be 100\%, since the ego-vehicle is possible to get caught in somewhere or drive to another direction with the target destination missing due to inaccurate control actions.

\subsubsection{Baselines} We implement two learning-based methods for comparisons: \texttt{NN-CIL} and \texttt{Set-CIL}. Network architectures are the same as our \texttt{G-CIL} except for the perceptual module. To illustrate the effectiveness of our graph representations, we change the perceptual module into MLP. 

Instead of GCN layers, \texttt{NN-CIL} utilizes 3-layer MLP and takes \textit{n-nearest} strategy to handle varying number of surrounding vehicles. The inputs are fixed 24-dimensional vector features, consisting of $\mathbf{x}^{ego}$ and three $\mathbf{x}_{i}$ with respect to the three nearest vehicles at time step $\mathit{t}$. Differently, \texttt{Set-CIL} takes 6-dimensional vectors as inputs. At each time step, $\mathbf{x}^{ego}$ and $\mathbf{x}_{i}$ of all surrounding vehicles are encoded sequentially with 3-layer MLP and summed up as the learned representation \cite{zaheer2017deep}. We train three methods with same training scenarios and take final convergence weights for performance evaluations and comparisons.

\subsection{Performance Evaluation}

\subsubsection{Quantitative results} To avoid the stochastic character, we test all methods 70 times for every scene and present final results in Table \ref{table:1}. It is obvious that our \texttt{G-CIL} outperforms the other two baselines, especially the generalization ability into unseen scenarios.

For \texttt{easy} setups, all three methods achieve high average success rate, among which average success rate of \texttt{NN-CIL} is the best as 73.81\%. Specifically, in \textit{Forward} task, \texttt{G-CIL} obtains the highest success rate 78.57\% while success rate of \texttt{Set-CIL} is the lowest as 35.71\%. However, the performance of \texttt{Set-CIL} in \textit{Turn Left} task is excellent with 100\% success rate. And it achieves the shortest navigation time in all tasks. Besides, although \texttt{NN-CIL} obtains the highest average success rate and lowest average collision rate, it is worth mentioning that only the ego-vehicle with policy learned from \texttt{NN-CIL} misses the target destination sometimes in \textit{Forward} task, caused by inaccurate steering angles. In such failure cases, generated steering angles are slightly but continuously smaller than correct values, which causes the ego-vehicle gradually shifts to the left lane and misses the target.  

For \texttt{middle} setups, average performance of all methods are worse. Our \texttt{G-CIL} maintains the similar level as 68.89\%, while success rates of the other two methods are lower than 50\%. Although navigation time of \texttt{Set-CIL} is also the shortest in both turning tasks, success rate is 0 in \textit{Forward} task. It seems that \texttt{Set-CIL} fails to learn the balance for different high-level commands, where performance of moving forward is the worst and turning left is the best; on the contrary, there is no such distinguish for \texttt{NN-CIL} and our \texttt{G-CIL}.

For \texttt{hard} setups, with the increasing of unseen moving types of surrounding vehicles, all average success rates further decrease to lower than 50\%, and our \texttt{G-CIL} is the best with 47.52\%. 
The performance of \texttt{Set-CIL} is quite unsatisfactory in terms of  \textit{Forward} task, as the ego-vehicle cannot move with small throttles. Our \texttt{G-CIL} achieves best performance on success rate, collision rate and average navigation time, which shows excellent robustness and generalization ability.

\subsubsection{Qualitative results} To further give the comprehensive analysis, we plot steering angle and throttle curves of three methods in successful trails for \textit{Turn Left} task, presented in Figure \ref{fig:curve}. It is shown that with the increasing complexity of the navigation environment, generated control actions tend to be shaky for all methods.

For successful cases, steering angle curves of three methods are very similar except for \texttt{Set-CIL}, which outputs steering angles larger than zero at early steps in \texttt{middle} and \texttt{hard} setups. It leads to the ego-vehicle driving offset to the right with shaky behaviors at first, although target destination is on the left. Besides, although success rates are different for \texttt{G-CIL} and \texttt{NN-CIL}, steering angle curves of success cases are both smooth.

On the contrary, throttle curves are quite distinguished, where \texttt{Set-CIL} is the most shaky one. Among three methods, control actions of \texttt{NN-CIL} is the smoothest, but the taken steps, that is, the navigation time is the largest. Throttle curves of our \texttt{G-CIL} tend to be shaky when approaching destinations. Specifically, the ego-vehicle slows down with smaller throttle vaules for \texttt{easy} and \texttt{middle} setups, while it accelerates for \texttt{hard} setup.

We encourage readers to view the additional video for a more detailed understanding of qualitative results, including success and failure cases of all methods. Although control action curves of \texttt{NN-CIL} seem to be more smooth, performance on action prediction accuracy, navigation time and driving speed are worse than \texttt{G-CIL}. Quantitative results, together with qualitative results, illustrate that our \texttt{G-CIL} learns the balance on effectiveness and robustness, with higher success rate and smooth driving behaviors.

\subsection{Ablation study}

When considering the dynamic scene of an unsignaled intersection with surrounding vehicles as graph-structure information, it is of great importance to define \textit{neighbors} of the ego-vehicle, that is, to define edges in the graph. In our \texttt{G-CIL}, we propose the strategy with weighted edges of n-closest vehicles and incomplete connectivity. To better demonstrate the performance of our proposed strategy, we further implement other edge definition strategies and evaluation results on \texttt{hard} setup with \textit{Forward} task are presented in Table \ref{table:2}. Every method is tested for 35 times.

As is shown, \textit{Fully-connection} refers to the fully connected graph without any weighted edges; \textit{Star-connection} means that no edges among surrounding vehicles and the ego-vehicle is linked with all surrounding vehicles in the scene with weights; \textit{Non-weighted} strategy is the unweighted version of our proposed method. Evaluations indicate that our strategy is most effective and efficient with the highest success rate and shorter navigation time. Moreover, with the comparison with \textit{Star-connection}, experimental results demonstrate that navigation policy performance improves with edges among surrounding vehicles, which helps to aggregate global environment information. Compared with the collision rate of \textit{Non-weighted}, it is shown empirically that weights corresponding to relative distances play an important role in collision avoidance.

\begin{center}
    \begin{table}[t]
        \renewcommand{\arraystretch}{1.5}
        \caption{Evaluation results for ablation study.}
        \label{table:2}
        \centering
        \begin{tabular}{c  c  c c}
        \toprule
        \multirow{1}{*}{Methods}&
        Success rate & Collision rate & Navigation time \\
        \midrule
        Ours  & \textbf{57.14\%}  & \textbf{42.86\%} & 15.45$s$ \\
        Fully-connection & 40.00\%  & 60.00\% & 15.95$s$\\
        Star-connection  & 45.71\%  & 54.29\% & 15.40$s$ \\
        Non-weighted  & 37.14\%  & 62.86\% & \textbf{14.41}$s$\\
        \bottomrule
        \end{tabular}
    \end{table}
    \vspace{-0.3cm}
\end{center}



\section{Conclusion}

In this paper, we propose a novel framework \texttt{G-CIL} for navigation through unsignaled intersections with varying number of surrounding vehicles and bicycles. We represent the dynamic traffic scenarios as graph-structured data and design an effective strategy to define the neighborhood relationship in the graph, to better capture and aggregate global environment information. On the basis of the combination of graph convolutional network (GCN) and conditional imitation learning (CIL), our proposed method achieves excellent performance and better generalization ability to unseen dynamic environments.

For the future work, we believe that incorporating road structure information into the perception module will make an improvement to the navigation policy performance, for the ability extension to handle more complicated dynamic environments. Besides, due to limitations of imitation learning algorithms, the learned policy performance can hardly outperforms the expert. So the combination of imitation learning and deep reinforcement learning is a promising direction for further researches.   










\bibliographystyle{IEEEtran}
\bibliography{ref.bib}

\end{document}